\documentclass{article}

\usepackage{times}
\usepackage{iclr2027_conference}
\usepackage{times}
\usepackage{amsmath,amssymb,mathtools}
\usepackage{centernot}
\usepackage{natbib}
\usepackage{booktabs,multirow}
\usepackage{graphicx}
\usepackage{xcolor}
\usepackage{microtype}
\usepackage{url}
\usepackage{hyperref}
\newcounter{algorithmbox}
\usepackage{multirow} 
\newenvironment{algorithmbox}[1]{%
\refstepcounter{algorithmbox}%
\par\medskip\noindent\begin{minipage}{\linewidth}\hrule\smallskip
\textbf{Algorithm \thealgorithmbox: #1}\par\smallskip
}{%
\smallskip\hrule\end{minipage}\par\medskip
}

\newcommand{\method}{\textbf{Capability-Aware Multi-Teacher On-Policy Distillation (CA-MOPD)}}

\title{Beyond Policy Discrepancy: Capability-Aware Multi-Teacher On-Policy Distillation}

\author{
Xuanfeng Yang \\
Tsinghua University \\
\texttt{yxf25@mails.tsinghua.edu.cn}
\And
Le Gai \\
Tsinghua University \\
\texttt{gl25@mails.tsinghua.edu.cn}
\And
Jiaming Wang \\
Tsinghua University \\
\texttt{wangjm25@mails.tsinghua.edu.cn}
\And
Xuan Zhang \\
Tsinghua University \\
\texttt{zhangx@cernet.edu.cn}
}

\iclrfinalcopy 
\begin{document}
\maketitle
\lhead{Preprint}

\begin{abstract}
Multi-teacher on-policy distillation (MOPD) integrates domain specialists into a shared student, but determining how strongly each teacher should supervise remains challenging.
Existing discrepancy-based allocation can assign strong supervision even when the student already matches or exceeds the teacher's task
performance, since behavioral mismatch does not necessarily imply a remaining capability deficit.
We propose Capability-Aware Multi-Teacher On-Policy Distillation (CA-MOPD), which combines specialist merging with capability-guided allocation to account for what the student has already learned.
CA-MOPD first merges reinforcement-learning specialists into a strong student initialization, then uses verifier-estimated teacher--student capability gaps to adjust domain-level distillation strength.
Supervision decreases as capability gaps close and increases when estimated student capability regresses, while dense token-level guidance remains provided by the teachers.
Across six benchmarks spanning mathematics, code generation, and instruction following, CA-MOPD achieves an average score of 32.22, outperforming naive MOPD, Open-MOPD, and the merged initialization by 0.86, 0.83, and 0.74 points, respectively.
Controlled comparisons further support the benefits of merged initialization and capability-gap guidance under matched training budgets.
\end{abstract}

\section{Introduction}

Reinforcement-learning (RL) post-training has become an effective way to develop language models with strong domain-specific capabilities, including mathematical reasoning, code generation, and instruction following~\citep{shao2024deepseekmath,pyatkin2026generalizing,gao2026open}. These capabilities are often learned by independently trained specialists, making it costly to deploy and maintain one model per domain. Multi-teacher on-policy distillation (MOPD) offers a natural alternative by using domain-specific teachers to supervise trajectories generated by a shared student, thereby integrating multiple specialist capabilities into a single model~\citep{agarwal2024policy,ma2026mopd}.

Parameter merging offers another approach to integrating specialist capabilities into a single model by combining their parameters~\citep{ilharco2022editing,yadav2023ties}. Parameter merging and MOPD are naturally complementary: merging can integrate specialist capabilities before distillation, potentially providing a stronger initialization for the student. However, the extent to which this initialization improves subsequent MOPD remains insufficiently understood. 

A key challenge in MOPD is how to allocate optimization across teachers. Open-MOPD addresses this with token-share balancing, discrepancy-based allocation, and reward refresh~\citep{gao2026open}, assigning larger weights to domains with greater teacher--student policy discrepancy. However, policy discrepancy measures behavioral mismatch rather than whether the teacher still has a task-level capability advantage. A student can remain distributionally different from a teacher while already matching or exceeding its task performance, especially after specialist merging.

Motivated by this distinction, we introduce \method{},  which first merges RL specialists with Task Arithmetic~\citep{ilharco2022editing} to provide a stronger student initialization for subsequent on-policy distillation and then allocates domain-level distillation strength according to verifier-estimated teacher--student capability gaps. Supervision decreases as the estimated gap closes, while a regression-aware retention mechanism increases it when previously acquired capability deteriorates. We retain token-share balancing to correct response-length imbalance.

\begin{figure}[htbp]
\centering
\includegraphics[width=\linewidth]{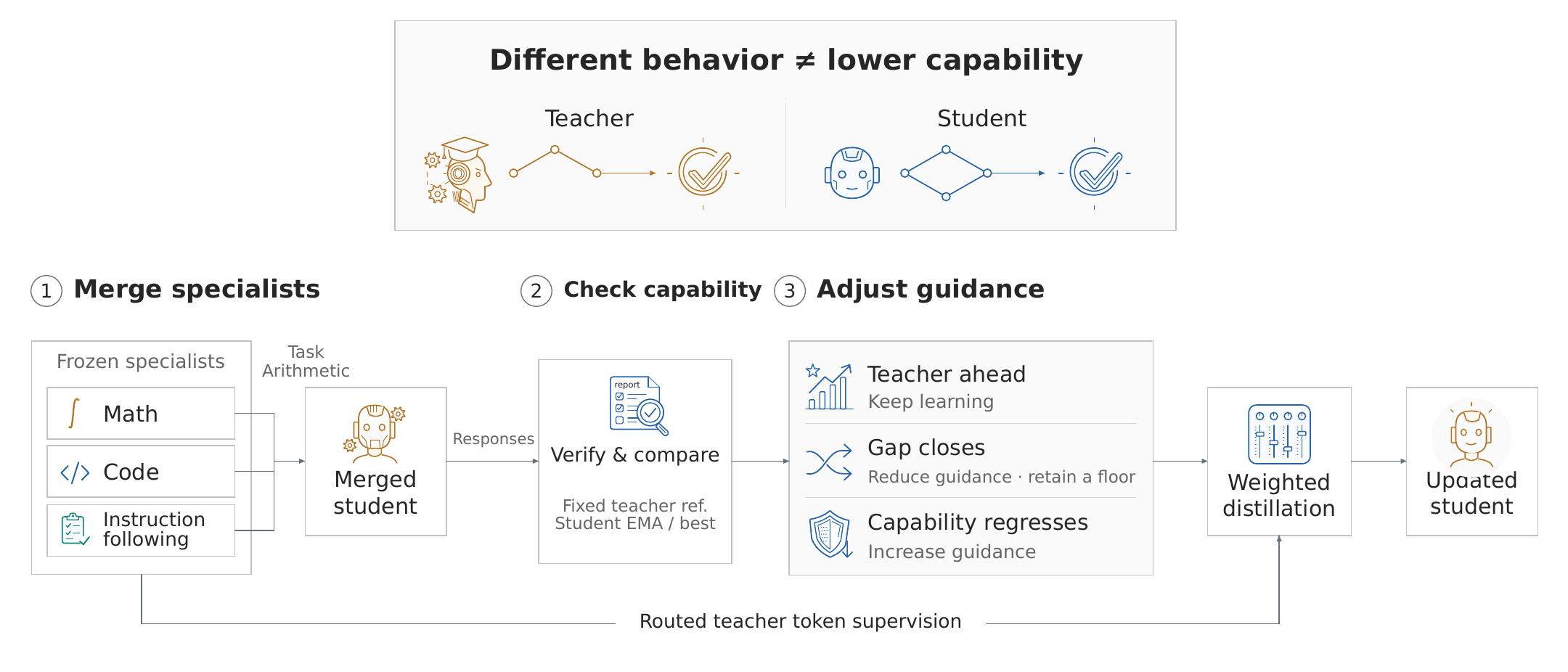}
\caption{Overview of CA-MOPD. (1) Domain RL specialists sharing a MixSFT base are merged into a strong student initialization. (2) Student responses are verified to track capability relative to a fixed, calibrated teacher reference and the student's historical best. (3) Remaining capability headroom and regression feedback control domain weights, which combine with token balancing and global scaling to modulate dense teacher supervision.}
\label{fig:ca_mopd_overview}
\end{figure}

Across six benchmarks spanning mathematics, code generation, and instruction following, CA-MOPD achieves an average score of $32.22$, compared with $31.36$ for naive MOPD, $31.39$ for Open-MOPD, and $31.48$ for the Task-Arithmetic initialization. The gains are concentrated in mathematics and instruction following, while code performance decreases relative to the merged initialization, reflecting a different cross-domain capability trade-off rather than uniform improvement.

Our main contributions are as follows:
\begin{itemize}
\item We show that initializing MOPD with merged RL specialists provides a stronger starting point than the shared SFT checkpoint and improves final aggregate performance under a fixed distillation budget.
\item We distinguish teacher--student policy discrepancy from signed capability gap, showing that behavioral mismatch does not necessarily indicate remaining task capability to transfer.
\item We propose \method{}, which combines specialist merging, verifier-based capability estimation, dynamic domain-level allocation, and regression-aware retention.
\item We show across three capability domains that CA-MOPD improves aggregate performance over naive MOPD, Open-MOPD, and the merged initialization, and analyze the resulting allocation dynamics.
\end{itemize}


\section{Background and Motivation}
\label{sec:background}

\subsection{Multi-Teacher On-Policy Distillation}
\label{sec:mopd_background}

Let $\mathcal{D}$ denote the domain set, $\pi_{\phi_d}$ the frozen specialist for domain $d$, and $\pi_\theta$ the shared student.
Given a prompt $x\sim\mathcal{P}_d$, the student samples
\begin{equation}
    y \sim \pi_{\theta_{\mathrm{old}}}(\cdot\mid x),
\end{equation}
and the domain teacher $\pi_{\phi_d}$ provides dense token-level supervision on this trajectory~\citep{ma2026mopd,gao2026open}.

The OPD signal $A_{itv}$ corresponds to candidate token $v$ at response position $t$ of example $i$.
The exact student Top-$K$ construction, candidate normalization, and clipped PPO objective are provided in Appendix~\ref{app:ca_objective}.

\subsection{Discrepancy-Based Distillation Allocation}
\label{sec:discrepancy_allocation}

Domains can differ in response length and convergence rate even under balanced prompt sampling.
Open-MOPD~\citep{gao2026open} addresses these differences through token-share balancing and discrepancy-based allocation.

Let $\mathcal{S}_{it}$ denote the candidate-token support used to compute $A_{itv}$, $M_{it}$ the valid-response mask, and
$N_d=\sum_{i:d(i)=d}\sum_t M_{it}$ the number of valid response tokens from domain $d$.
For the mean statistic used in our implementation, the domain discrepancy is
\begin{equation}
    m_d
    =
    \frac{1}{N_d}
    \sum_{i:d(i)=d}\sum_t
    M_{it}
    \left|
    \sum_{v\in\mathcal{S}_{it}} A_{itv}
    \right|.
    \label{eq:open_gap}
\end{equation}
Larger $m_d$ indicates a larger token-level mismatch and therefore receives stronger supervision under discrepancy-following allocation.

\subsection{Policy Discrepancy Is Not Capability Deficit}
\label{sec:capability_gap}

Let $V_d(x,y)\in[0,1]$ be the domain verifier score for response $y$ to prompt $x$.
We define policy capability as
\begin{equation}
    J_d(\pi)
    =
    \mathbb{E}_{x\sim\mathcal{P}_d,\,
    y\sim\pi(\cdot\mid x)}
    \left[V_d(x,y)\right],
    \label{eq:capability}
\end{equation}
and the signed teacher--student capability gap as
\begin{equation}
    g_d(\theta)
    =
    J_d(\pi_{\phi_d})
    -
    J_d(\pi_\theta).
    \label{eq:signed_gap}
\end{equation}
Thus, $g_d>0$ indicates a teacher capability advantage, while $g_d\leq 0$ indicates that the student matches or exceeds the teacher under the same domain distribution and verifier.

Policy discrepancy $m_d$ measures behavioral mismatch without establishing a teacher capability advantage:
\begin{equation}
    m_d>0
    \quad\centernot\Rightarrow\quad
    g_d(\theta)>0.
    \label{eq:discrepancy_not_capability}
\end{equation}
This distinction is especially relevant after specialist merging, which can transfer task capabilities while leaving policy discrepancy nonzero.

\section{Capability-Aware Multi-Teacher On-Policy Distillation}
\label{sec:method}

Figure~\ref{fig:ca_mopd_overview} summarizes the CA-MOPD workflow.
CA-MOPD implements the preceding principle in four steps.
It first merges independently trained RL specialists into a strong multi-domain initialization, calibrates each teacher's task-level capability, tracks the student's corresponding capability from verifier-scored on-policy rollouts, and converts the remaining capability gap into a domain-level distillation factor.
This factor is then combined with token-share balancing and applied to the common dense OPD objective.

\subsection{Merged Initialization}
\label{sec:merge}

We assume that all domain specialists are obtained by applying domain-specific RL to the same mixed-domain SFT checkpoint with parameters $\theta_0$.
Let $\phi_d$ denote the parameters of specialist $d$.
Following Task Arithmetic~\citep{ilharco2022editing}, we initialize the student as
\begin{equation}
    \theta_{\mathrm{merge}}
    =
    \theta_0
    +
    \sum_{d\in\mathcal{D}}
    \alpha_d(\phi_d-\theta_0),
    \qquad
    \alpha_d=1.
    \label{eq:merge}
\end{equation}

The merge transfers part of the specialist updates before on-policy distillation and therefore changes the residual capability profile that distillation must address.
CA-MOPD treats $\theta_{\mathrm{merge}}$ as the starting point and allocates subsequent supervision according to the capability that remains to be transferred rather than assuming that every teacher retains a positive task-level advantage.

\subsection{Verifier-Calibrated Capability Tracking}
\label{sec:capability_estimation}

Following Eq.~\eqref{eq:capability}, we estimate each teacher's capability once on a held-out calibration set $\mathcal{C}_d$:
\begin{equation}
    T_d
    =
    \widehat J_{\mathcal C_d}(\pi_{\phi_d}).
    \label{eq:teacher_reference}
\end{equation}
The merged student is evaluated with the same protocol to initialize
\begin{equation}
    C_d^{(0)}
    =
    \widehat J_{\mathcal C_d}(\pi_{\theta_{\mathrm{merge}}}),
    \qquad
    C_d^{\star(0)}=C_d^{(0)},
    \qquad
    H_d^{(0)}=[T_d-C_d^{(0)}]_+,
    \label{eq:initial_capability}
\end{equation}
where $[z]_+=\max(z,0)$ and $C_d^{\star}$ records the best student capability estimate observed so far.
The calibration split, sampling protocol, and score aggregation are detailed in Appendix~\ref{app:calibration}.

During training, CA-MOPD reuses verifier outcomes from the student's existing on-policy rollouts rather than generating an additional rollout for capability estimation.
Let $Z_d^{(k)}$ denote the domain-level verifier observation at controller update $k$.
We maintain an exponential moving average
\begin{equation}
    C_d^{(k)}
    =
    \beta C_d^{(k-1)}
    +
    (1-\beta)Z_d^{(k)},
    \label{eq:capability_ema}
\end{equation}
and update the historical best as
\begin{equation}
    C_d^{\star(k)}
    =
    \max
    \left\{
    C_d^{\star(k-1)},
    C_d^{(k)}
    \right\}.
    \label{eq:historical_best}
\end{equation}
The online signed gap and positive headroom are then
\begin{equation}
    g_d^{(k)}
    =
    T_d-C_d^{(k)},
    \qquad
    H_d^{(k)}
    =
    [g_d^{(k)}]_+.
    \label{eq:online_capability}
\end{equation}
Thus, $T_d$ provides a fixed teacher reference, while $C_d^{(k)}$ tracks the student's current capability under the same verifier family.

\subsection{Capability-Aware Distillation Controller}
\label{sec:capability_controller}

The controller captures two signals: the amount of teacher advantage that remains and whether previously acquired student capability has regressed.

\paragraph{Remaining capability headroom.}
For domains with non-negligible initial headroom, we measure the fraction of the original deficit that remains:
\begin{equation}
    r_d^{(k)}
    =
    \frac{H_d^{(k)}}{H_d^{(0)}}.
    \label{eq:headroom_ratio_main}
\end{equation}
The corresponding headroom factor is
\begin{equation}
    f_{d,\mathrm{head}}^{(k)}
    =
    \operatorname{clip}
    \left(
    \left(r_d^{(k)}\right)^p,
    f_{\min},
    f_{\max}
    \right).
    \label{eq:head_factor}
\end{equation}
As the estimated teacher advantage closes, this factor decreases; a positive floor keeps teacher feedback available instead of permanently disabling the domain.
Numerical handling of near-zero initial headroom and all controller hyperparameters are given in Appendix~\ref{app:ca_controller}.

\paragraph{Regression-aware retention.}
Because continued optimization on other domains can reduce a previously acquired capability, we compare the current estimate with its historical best:
\begin{equation}
    R_d^{(k)}
    =
    \left[
    C_d^{\star(k)}
    -
    C_d^{(k)}
    -
    \tau
    \right]_+.
    \label{eq:regression}
\end{equation}
When $R_d^{(k)}>0$, a bounded guard factor
$f_{d,\mathrm{guard}}^{(k)}$ increases supervision according to the detected regression.
The final capability factor is
\begin{equation}
    f_d^{(k)}
    =
    \begin{cases}
    \max\!\left\{
    f_{d,\mathrm{head}}^{(k)},
    f_{d,\mathrm{guard}}^{(k)}
    \right\},
    & R_d^{(k)}>0,\\
    f_{d,\mathrm{head}}^{(k)},
    & R_d^{(k)}=0.
    \end{cases}
    \label{eq:capability_factor}
\end{equation}
The exact guard mapping, warm-up, and controller constants are provided in Appendix~\ref{app:ca_controller}.
The headroom term therefore allocates supervision toward capability that remains to be transferred, while the retention term responds when previously observed capability declines.

\subsection{Capability-Weighted OPD}
\label{sec:weighted_objective}

Capability allocation does not replace token-share balancing: the former estimates how much task-level transfer remains useful, whereas the latter corrects unequal token contributions caused by response-length differences.
Let $s_d^{(k)}$ be the observed response-token share of domain $d$ and $q_d$ its target share.
We retain the Open-MOPD balancing factor
\begin{equation}
    b_d^{(k)}
    =
    \frac{q_d}{s_d^{(k)}}.
    \label{eq:token_balance_method}
\end{equation}
The capability factor and token-balancing factor are combined into the actual per-token multiplier
\begin{equation}
    w_d^{(k)}
    =
    F^{(k)}
    \frac{
    b_d^{(k)} f_d^{(k)}
    }{
    \sum_j
    s_j^{(k)}
    b_j^{(k)}
    f_j^{(k)}
    },
    \qquad
    F^{(k)}
    =
    \max_j f_j^{(k)}.
    \label{eq:final_weight}
\end{equation}

For example $i$ from domain $d(i)$, CA-MOPD scales the common dense OPD signal as
\begin{equation}
    \widehat A_{itv}
    =
    \operatorname{sg}
    \left[
    w_{d(i)}^{(k)} A_{itv}
    \right].
    \label{eq:weighted_advantage}
\end{equation}
The resulting weighted signal is optimized with the token-normalized PPO objective~\citep{schulman2017proximal} used by the Open-MOPD implementation~\citep{gao2026open}.
The full candidate-level OPD signal, clipping rule, and optimization objective are given in Appendix~\ref{app:ca_objective}.

Overall, CA-MOPD leaves the token-level teacher supervision mechanism unchanged and modifies only how strongly each domain contributes to optimization.
The teacher provides the local distillation direction, while verifier-estimated capability feedback determines the domain-level optimization pressure.

\section{Experiments}
\label{sec:experiments}


\subsection{Experimental Setup}
\label{sec:exp_setup}

We use SmolLM3-3B~\citep{bakouch2025smollm3} with the MixSFT checkpoint and three RL specialists released by Open-MOPD~\citep{gao2026open}. The specialists cover mathematics, code generation, and instruction following and share the same MixSFT initialization.
The main CA-MOPD model is initialized from the Task-Arithmetic merge of the three specialists and trained for 200 outer steps.
We evaluate mathematics on AIME24 and AIME25, code generation on LiveCodeBench v5 and v6~\citep{jain2025livecodebench}, and instruction following on IFEval~\citep{zhou2023instruction} and IFBench~\citep{pyatkin2026generalizing}.
All six benchmark scores are reported as percentages, and their equal-weight mean is used as the aggregate score.
Unless otherwise stated, comparisons use the same training budget and evaluation protocol.
Full training, decoding, and benchmark configurations are provided in Table~\ref{tab:hyperparameters} and Appendix~\ref{app:ca_eval}.

\subsection{Overall Multi-Domain Performance}
\label{sec:overall_results}

Table~\ref{tab:main} shows that CA-MOPD achieves the highest six-benchmark average of $32.22$, exceeding ParamMerge-TA, Naive MOPD, and Open-MOPD by $0.74$, $0.86$, and $0.83$ points, respectively.
It ranks first on both mathematics benchmarks and IFBench, with gains of $2.19$ points on AIME24 and $3.34$ points on IFBench over Open-MOPD.
These results demonstrate the effectiveness of CA-MOPD in integrating specialist capabilities into a single model, achieving stronger aggregate performance than parameter merging alone and existing MOPD baselines.

\begin{table}[t]
\centering
\small
\setlength{\tabcolsep}{2.7pt}
\caption{
Overall performance across mathematics, code generation, and instruction following.
All values are percentages.
\textbf{Bold} indicates the highest score in each column among all single-model methods.
}
\label{tab:main}
\begin{tabular}{lrrrrrrr}
\toprule
Method & AIME24 & AIME25 & LCBv5 & LCBv6 & IFEval & IFBench & Avg \\
\midrule
MixSFT
& 14.58 & 17.14 & 14.49 & 19.89 & 68.76 & 14.67 & 24.92 \\
RouteRL
& 23.23 & 24.27 & 23.11 & 22.06 & 71.90 & 22.00 & 31.10 \\
\midrule
ParamMerge-TA
& 23.80 & 23.23 & \textbf{23.83} & \textbf{22.69} & 73.01 & 22.33 & 31.48 \\
Naive MOPD
& 22.92 & 24.79 & 21.92 & 22.34 & 74.86 & 21.33 & 31.36 \\
Open-MOPD
& 22.76 & 25.00 & 22.99 & 21.66 & \textbf{75.60} & 20.33 & 31.39 \\
CA-MOPD
& \textbf{24.95} & \textbf{25.10} & 22.75 & 21.60 & 75.23 & \textbf{23.67} & \textbf{32.22} \\
\bottomrule
\end{tabular}
\end{table}

\subsection{Specialist Merging Provides a Strong Initialization for MOPD}
\label{sec:init_ablation}

We next examine whether specialist merging provides a stronger starting point for MOPD. To isolate the effect of initialization, we train CA-MOPD for the same 200-step budget from either MixSFT or ParamMerge-TA. Table~\ref{tab:init_ablation} reports performance before and after CA-MOPD training.

\begin{table}[t]
\centering
\small
\setlength{\tabcolsep}{2.7pt}
\caption{
Effect of initialization under a fixed 200-step CA-MOPD budget.
We report performance before and after training; \textbf{bold} marks the better post-training result for each metric.
}
\label{tab:init_ablation}
\begin{tabular}{llrrrrrrr}
\toprule
Initialization & Stage
& AIME24 & AIME25 & LCBv5 & LCBv6 & IFEval & IFBench & Avg \\
\midrule
\multirow{2}{*}{MixSFT}
& Initial
& 14.58 & 17.14 & 14.49 & 19.89
& 68.76 & 14.67 & 24.92 \\
& + CA-MOPD
& 21.93 & \textbf{25.52} & \textbf{23.11} & \textbf{23.31}
& 72.64 & 22.67 & 31.53 \\
\midrule
\multirow{2}{*}{ParamMerge-TA}
& Initial
& 23.80 & 23.23 & 23.83 & 22.69
& 73.01 & 22.33 & 31.48 \\
& + CA-MOPD
& \textbf{24.95} & 25.10 & 22.75 & 21.60
& \textbf{75.23} & \textbf{23.67} & \textbf{32.22} \\
\bottomrule
\end{tabular}
\end{table}

ParamMerge-TA already provides a strong multi-domain initialization, improving the aggregate score from $24.92$ to $31.48$ before any on-policy distillation. Notably, this nearly matches the $31.53$ obtained after training CA-MOPD from MixSFT for the full 200-step budget, suggesting that specialist merging can integrate a substantial portion of the target capabilities before MOPD begins.

With the same subsequent 200-step CA-MOPD budget, the merged initialization further improves the final average to $32.22$. Although the gains are not uniform across domains, this result shows that merging and MOPD are complementary: merging provides a stronger capability-integrated starting point, while subsequent on-policy distillation refines the remaining capability gaps.

\subsection{Ablation of Capability-Gap Guidance}
\label{sec:gap_ablation}

\begin{figure}[t]
\centering
\includegraphics[width=\linewidth]{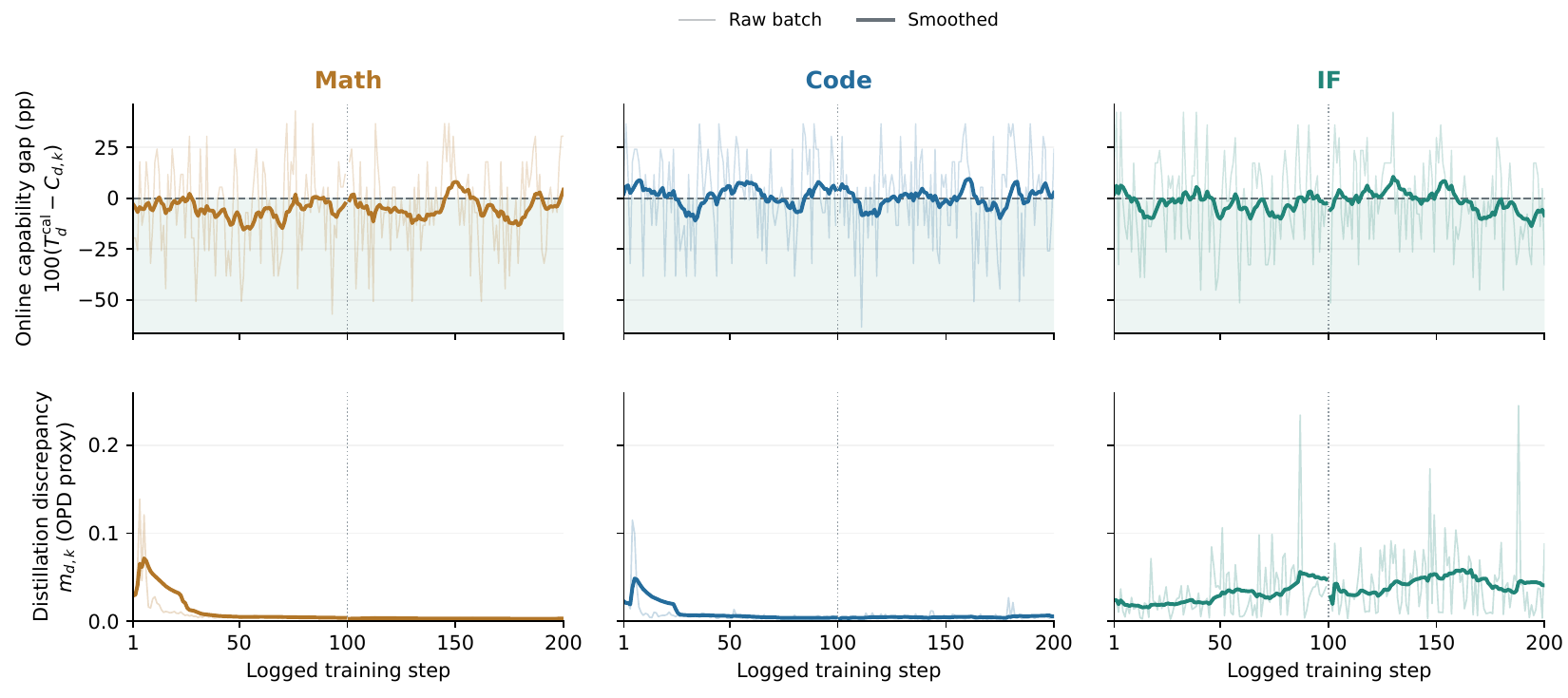}
\caption{Capability-gap and discrepancy dynamics during 200 steps of ParamMerge-initialized CA-MOPD. Top: fixed calibrated teacher score minus student verifier score, in percentage points; faint lines use raw batch scores and bold lines use the logged EMA. Bottom: the OPD discrepancy proxy, shown as raw values and a trailing 20-step mean. Negative online gaps indicate that the student estimate exceeds the calibrated teacher reference. These are training-time diagnostics on changing prompts, not fixed-benchmark capability trajectories; IF uses the \texttt{content\_ok} verifier.}
\label{fig:gap_curves}
\end{figure}

We first examine whether capability gap and policy discrepancy provide equivalent signals for allocating distillation effort. Figure~\ref{fig:gap_curves} shows that they exhibit substantially different training dynamics. In particular, a nonpositive EMA capability gap coexists with positive policy discrepancy at 170, 80, and 129 of the 200 recorded steps for mathematics, code, and instruction following, respectively. This behavior is especially common because the ParamMerge-TA initialization already approaches or exceeds the calibrated teacher capability in several domains, while substantial token-level discrepancy remains. Thus, policy discrepancy can persist even when little or no positive teacher capability advantage remains.

We next compare the step-50 CA-MOPD checkpoint with a distribution-discrepancy variant, both originating from MixSFT.
The variant replaces the capability-headroom ratio with an EMA-smoothed discrepancy $m_d$ from Eq.~\eqref{eq:open_gap}, normalized by its first-batch value, while retaining the capability-based regression guard, token balancing, warm-up, and clipping.

\begin{table}[t]
\centering
\small
\setlength{\tabcolsep}{3pt}
\caption{
Step-50 comparison of capability-gap and distribution-discrepancy guidance from MixSFT.
Both variants retain the capability-based regression guard.
Scores are percentages and Avg is the equal-weight six-benchmark mean.
\textbf{Bold} marks the higher score in each column.
}
\label{tab:gap_ablation}
\begin{tabular}{lrrrrrrr}
\toprule
Allocation signal & AIME24 & AIME25 & LCBv5 & LCBv6 & IFEval & IFBench & Avg \\
\midrule
Capability gap
& 21.98 & 25.00 & \textbf{21.74} & \textbf{22.46} & \textbf{74.12} & \textbf{23.00} & \textbf{31.38} \\
Distribution discrepancy
& \textbf{24.06} & \textbf{25.73} & 20.54 & 22.17 & 73.01 & 21.00 & 31.09 \\
\bottomrule
\end{tabular}
\end{table}

Table~\ref{tab:gap_ablation} shows that capability-gap guidance achieves a higher six-benchmark average ($31.38$ versus $31.09$).
It improves LCBv5, LCBv6, IFEval, and IFBench by $1.20$, $0.29$, $1.11$, and $2.00$ points, respectively, while distribution-discrepancy guidance performs better on both mathematics benchmarks.
Together with the distinct signal dynamics in Figure~\ref{fig:gap_curves}, these results support the usefulness of capability-gap guidance for aggregate multi-domain performance, with gains in code generation and instruction following in this comparison.

\subsection{Training Allocation Dynamics}
\label{sec:dynamics}

\begin{figure}[t]
\centering
\includegraphics[width=\linewidth]{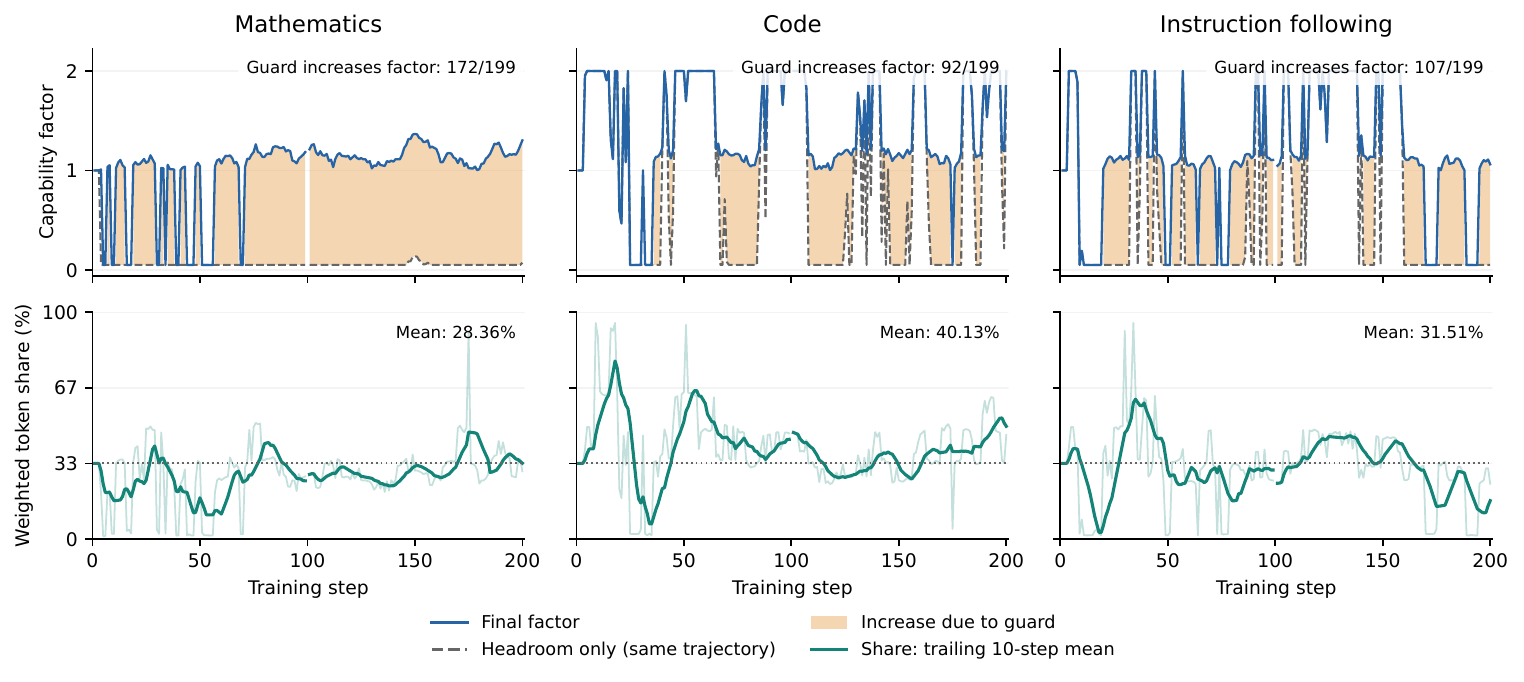}
\caption{Training allocation during ParamMerge-initialized CA-MOPD.
Top: logged final capability factors (blue) and reconstructed headroom-only factors on the same trajectory (dashed gray); orange shading marks increases due to the regression guard. Both factors include the same warm-up override. Count denominators indicate the 200-step training budget.
Bottom: weighted token shares, with raw values in faint lines and trailing
10-step means in bold; dotted lines indicate equal shares.
These are controller diagnostics on changing training batches, not
held-out capability curves.}
\label{fig:controller_dynamics}
\end{figure}

Figure~\ref{fig:controller_dynamics} shows how capability feedback modulates training allocation in the ParamMerge-initialized run.
We report the weighted token share
$a_d^{(k)}=s_d^{(k)}w_d^{(k)}/\sum_j s_j^{(k)}w_j^{(k)}$,
which summarizes relative token weighting after response-length correction, but should not be interpreted as measured gradient contribution.
Table~\ref{tab:allocation_outcomes} relates these shares and regression-guard activity to domain-level performance changes over 200 training steps.

To isolate the guard's effect on the controller output, we reconstruct the headroom-only factor on the same recorded capability trajectory, preserving clipping and warm-up.
The guard increases the final factor at 172, 92, and 107 of the 200 recorded steps for mathematics, code, and instruction following, respectively.
This effect is most pronounced in mathematics, where the initially nonpositive teacher advantage would otherwise keep the headroom-only factor near its floor.
Under the full controller, mathematics improves from $23.52$ to $25.03$, while instruction following improves from $47.67$ to $49.45$.
These trajectories are consistent with the guard retaining useful supervision even when the estimated teacher advantage is small.

Allocation magnitude, however, does not directly predict downstream gain.
Code receives the largest average weighted token share ($40.13\%$) yet decreases by $1.08$ points, reflecting domain-specific factors such as teacher-signal quality, cross-domain interference, and optimization dynamics.
We therefore interpret allocation as a domain-conditioned response to the estimated capability state rather than as a universal predictor of downstream improvement.
Importantly, the observed trajectories confirm that CA-MOPD produces substantially different supervision patterns across domains in response to their evolving capability states, which is precisely the behavior targeted by the proposed adaptive allocation mechanism.

\begin{table}[t]
\centering
\small
\setlength{\tabcolsep}{5pt}
\caption{Training allocation and domain-level outcomes for ParamMerge-initialized
CA-MOPD. Share is the mean weighted token share over 200 recorded steps;
guard steps count actual factor increases above the reconstructed headroom-only
reference. Domain scores average the two corresponding benchmarks in
Table~\ref{tab:main}; changes are computed before rounding, in percentage
points. This is a descriptive comparison, not a component ablation.}
\label{tab:allocation_outcomes}
\begin{tabular}{lrrrrr}
\toprule
Domain & Share (\%) & Guard steps & Initial & Final & $\Delta$ \\
\midrule
Mathematics & 28.36 & 172/200 & 23.52 & 25.03 & $+1.51$ \\
Code & 40.13 & 92/200 & 23.26 & 22.18 & $-1.08$ \\
Instruction following & 31.51 & 107/200 & 47.67 & 49.45 & $+1.78$ \\
\bottomrule
\end{tabular}
\end{table}

\subsection{Computational Cost}
\label{sec:cost}

CA-MOPD estimates online capability from verifier outcomes on existing student rollouts and uses these estimates to reweight domain-level supervision, without requiring additional rollout generation.
Table~\ref{tab:cost} reports the mean per-step runtime of the main training components over the 200 steps with available timing records.
Student rollout generation and actor optimization together account for $83.97\%$ of the measured step time, while model forward passes for computing token probabilities used in distillation account for $12.62\%$.
Online capability feedback takes $33.30$ seconds per step, corresponding to $3.20\%$ of the measured step time.
These results indicate that online capability feedback constitutes a small fraction of the training runtime in this setting.

\begin{table}[t]
\centering
\small
\caption{Runtime breakdown of CA-MOPD training, averaged over the 200 steps with available timing records.
Shares are relative to the total measured step time; the listed components account for $99.79\%$ of this total.}
\label{tab:cost}
\begin{tabular}{lrr}
\toprule
Component & Mean time (s/step) & Share (\%) \\
\midrule
Student rollout generation & 588.22 & 56.56 \\
Model forward passes & 131.19 & 12.62 \\
Actor optimization & 285.03 & 27.41 \\
Online capability feedback & 33.30 & 3.20 \\
\bottomrule
\end{tabular}
\end{table}

\section{Related Work}
\label{sec:related_work}

\paragraph{On-policy and multi-teacher distillation.}
Knowledge distillation transfers teacher behavior to a student \citep{hinton2015distilling,kim2016sequence}, while GKD extends distillation to student-generated trajectories \citep{agarwal2024policy}. MOPD further integrates multiple domain-specialized RL teachers into a shared student \citep{ma2026mopd}, with recent variants exploring richer teacher organizations \citep{wang2026mad,yin2026h}. Our work focuses instead on how strongly each teacher should supervise once specialist capabilities have been partially integrated.

\paragraph{Adaptive allocation in multi-teacher distillation.}
Open-MOPD combines token-share balancing with discrepancy-based allocation and reward refresh \citep{gao2026open}, while D$^3$-MOPD adapts training allocation using distillation trajectories and convergence dynamics \citep{sun2026d}. These approaches primarily infer optimization demand from distillation-side signals. SPOT uses verifier-scored student continuations to calibrate token-level distillation targets~\citep{qu2026spot}. CA-MOPD uses verifier outcomes to estimate signed domain-level capability gaps and adjust distillation strength while retaining the teacher-provided token-level signal.

\paragraph{Model merging.}
Weight-space merging integrates independently fine-tuned models without increasing inference-time model count, including Model Soups \citep{wortsman2022model}, Task Arithmetic \citep{ilharco2022editing}, and TIES-Merging \citep{yadav2023ties}. Rather than introducing a new merging operator, we use Task Arithmetic to obtain a strong initialization and study how subsequent on-policy distillation can refine the remaining capability gaps.

\section{Limitations}

CA-MOPD depends on verifier scores to estimate capability, so its allocation quality is limited by the calibration distribution and verification procedure, while domain-level averages may hide example-level differences. The current controller also relies on smoothed point estimates and historical-best tracking rather than calibrated uncertainty, making small capability gaps potentially sensitive to noise. In addition, CA-MOPD continuously reweights domains but does not skip teacher computation, so it provides neither hard non-forgetting guarantees nor automatic compute savings. Finally, our evaluation is limited to one 3B model family, three domains, and known domain routing; broader generalization to larger models, heterogeneous teachers, more domains, and learned routing remains open.

\section{Conclusion}

In this work, we presented CA-MOPD, which combines specialist merging with capability-aware multi-teacher on-policy distillation. Our experiments show that initializing CA-MOPD with merged RL specialists provides a stronger starting point and improves final aggregate performance over initialization from the shared SFT checkpoint under the same training budget. Building on this initialization, CA-MOPD uses verifier-estimated teacher--student capability gaps and student regression signals to adapt domain-level supervision, accounting for capabilities already integrated into the student. Across six benchmarks spanning mathematics, code generation, and instruction following, the resulting approach achieves higher aggregate performance than naive MOPD, Open-MOPD, and the Task-Arithmetic merged initialization, although gains are not uniform across domains. These findings support combining parameter merging with subsequent on-policy distillation and highlight the importance of considering both the student's initial capabilities and the allocation of teacher supervision.

\section*{AI Use Statement}

We used generative AI tools to assist with refining research hypotheses, providing feedback on methodology and experimental design, interpreting intermediate results, inspecting code and configurations, organizing literature, and drafting and editing the manuscript.
All AI-assisted content was independently reviewed and verified by the authors, including citations, technical claims, code, configurations, and experimental results. The authors made all final research decisions and take full responsibility for the content of this work.

\section*{Ethics Statement}

This work studies the integration of publicly available language-model checkpoints on mathematics, code, and instruction-following tasks. It does not involve human-subject experiments or the collection of private personal data. The resulting methods may inherit harmful behaviors or biases from their base models, teachers, training data, and verifiers. We therefore do not claim that improved benchmark integration implies broader safety or reliability, and recommend task-specific safety evaluation before deployment.

\section*{Reproducibility Statement}

Appendix~\ref{app:method_details} specifies the capability controller, its hyperparameters, the capability-weighted OPD objective, and the training algorithm.
Appendix~\ref{app:implementation} and Table~\ref{tab:hyperparameters} document the model family, merged initialization, hardware, training framework, sampling and optimization settings, training budget, and checkpoint schedule, together with the controller state and other training states required to resume training.
Appendix~\ref{app:calibration} describes the held-out calibration sets of 256 prompts per domain, four responses per prompt, capability-score aggregation, and reuse of training rollouts for online capability estimation.
Appendix~\ref{app:ca_eval} and Table~\ref{tab:eval_protocol} specify the six evaluation benchmarks, decoding settings, and scoring protocols, including average sampled accuracy for mathematics and code generation and strict prompt-level correctness for instruction following.
\bibliography{iclr2027_conference}

@inproceedings{agarwal2024policy,
  title     = {On-policy distillation of language models: Learning from self-generated mistakes},
  author    = {Agarwal, Rishabh and Vieillard, Nino and Zhou, Yongchao and Stanczyk, Piotr and Ramos Garea, Sabela and Geist, Matthieu and Bachem, Olivier},
  booktitle = {International Conference on Learning Representations},
  volume    = {2024},
  pages     = {21246--21263},
  year      = {2024}
}

@article{ma2026mopd,
  title   = {Mopd: Multi-teacher on-policy distillation for capability integration in llm post-training},
  author  = {Ma, Wenhan and Wei, Jianyu and Zhao, Liang and Zhang, Hailin and Xiao, Bangjun and Li, Lei and Yang, Qibin and Gao, Bofei and Wang, Yudong and Li, Rang and others},
  journal = {arXiv preprint arXiv:2606.30406},
  year    = {2026}
}

@article{gao2026open,
  title   = {Open-MOPD: Diagnosing and Fixing Capability Imbalance in Multi-Teacher On-Policy Distillation},
  author  = {Gao, Huan-ang and Chi, Haohan and Yan, Yong and Feng, Shiyuan and Wu, Hanlin and Jiang, Zheng and He, Bingxiang and Ma, Wei-Ying and Zhang, Ya-Qin and Zhou, Hao},
  journal = {arXiv preprint arXiv:2608.19098},
  year    = {2026}
}

@article{qu2026spot,
  title   = {SPOT: Sparse Probing and Outcome Calibration for On-Policy Distillation},
  author  = {Qu, Zikun and Zhang, Min and Kong, Mingze and Shang, Zhiwei and Ban, Yikun and Qiu, Shuang and Dai, Zhongxiang},
  journal = {arXiv preprint arXiv:2608.04419},
  year    = {2026}
}

@article{wang2026mad,
  title   = {Mad-opd: Breaking the ceiling in on-policy distillation via multi-agent debate},
  author  = {Wang, Jianze and Liu, Ying and Chen, Jinlong and Hu, Xuchun and Zhang, Qilong and Cao, Yu and Wang, Jun and Yang, Hua and Xie, Yong and Chen, Qianglong},
  journal = {arXiv preprint arXiv:2605.01347},
  year    = {2026}
}

@article{ilharco2022editing,
  title   = {Editing models with task arithmetic},
  author  = {Ilharco, Gabriel and Ribeiro, Marco Tulio and Wortsman, Mitchell and Gururangan, Suchin and Schmidt, Ludwig and Hajishirzi, Hannaneh and Farhadi, Ali},
  journal = {arXiv preprint arXiv:2212.04089},
  year    = {2022}
}

@article{yadav2023ties,
  title   = {Ties-merging: Resolving interference when merging models},
  author  = {Yadav, Prateek and Tam, Derek and Choshen, Leshem and Raffel, Colin A and Bansal, Mohit},
  journal = {Advances in neural information processing systems},
  volume  = {36},
  pages   = {7093--7115},
  year    = {2023}
}

@article{shao2024deepseekmath,
  title   = {Deepseekmath: Pushing the limits of mathematical reasoning in open language models},
  author  = {Shao, Zhihong and Wang, Peiyi and Zhu, Qihao and Xu, Runxin and Song, Junxiao and Bi, Xiao and Zhang, Haowei and Zhang, Mingchuan and Li, YK and Wu, Yang and others},
  journal = {arXiv preprint arXiv:2402.03300},
  year    = {2024}
}

@article{hinton2015distilling,
  title   = {Distilling the knowledge in a neural network},
  author  = {Hinton, Geoffrey and Vinyals, Oriol and Dean, Jeff},
  journal = {arXiv preprint arXiv:1503.02531},
  year    = {2015}
}

@article{schulman2017proximal,
  title   = {Proximal policy optimization algorithms},
  author  = {Schulman, John and Wolski, Filip and Dhariwal, Prafulla and Radford, Alec and Klimov, Oleg},
  journal = {arXiv preprint arXiv:1707.06347},
  year    = {2017}
}

@inproceedings{ye2020mastering,
  title     = {Mastering complex control in moba games with deep reinforcement learning},
  author    = {Ye, Deheng and Liu, Zhao and Sun, Mingfei and Shi, Bei and Zhao, Peilin and Wu, Hao and Yu, Hongsheng and Yang, Shaojie and Wu, Xipeng and Guo, Qingwei and others},
  booktitle = {Proceedings of the AAAI conference on artificial intelligence},
  volume    = {34},
  number    = {04},
  pages     = {6672--6679},
  year      = {2020}
}

@inproceedings{jain2025livecodebench,
  title     = {Livecodebench: Holistic and contamination free evaluation of large language models for code},
  author    = {Jain, Naman and Gu, Alex and Li, Wen-Ding and Yan, Fanjia and Zhang, Tianjun and Wang, Sida and Solar-Lezama, Armando and Sen, Koushik and Stoica, Ion},
  booktitle = {International Conference on Learning Representations},
  volume    = {2025},
  pages     = {58791--58831},
  year      = {2025}
}

@article{zhou2023instruction,
  title   = {Instruction-following evaluation for large language models},
  author  = {Zhou, Jeffrey and Lu, Tianjian and Mishra, Swaroop and Brahma, Siddhartha and Basu, Sujoy and Luan, Yi and Zhou, Denny and Hou, Le},
  journal = {arXiv preprint arXiv:2311.07911},
  year    = {2023}
}

@article{pyatkin2026generalizing,
  title   = {Generalizing verifiable instruction following},
  author  = {Pyatkin, Valentina and Malik, Saumya and Graf, Victoria and Ivison, Hamish and Huang, Shengyi and Dasigi, Pradeep and Lambert, Nathan and Hajishirzi, Hanna},
  journal = {Advances in Neural Information Processing Systems},
  volume  = {38},
  year    = {2026}
}

@article{sun2026d,
  title   = {D $^3$-MOPD: Adaptive Dynamic Domain ScheDuling for Efficient Multi-Teacher Distillation},
  author  = {Sun, Zechen and Zhang, Zhiwei and Zhao, Fei and Li, Juntao and Chuan, Mu and Deng, Huayu and Zhan, Guojian and Chen, Wenliang and Hu, Yao and Zhang, Min},
  journal = {arXiv preprint arXiv:2608.24987},
  year    = {2026}
}

@inproceedings{kim2016sequence,
  title     = {Sequence-level knowledge distillation},
  author    = {Kim, Yoon and Rush, Alexander M},
  booktitle = {Proceedings of the 2016 conference on empirical methods in natural language processing},
  pages     = {1317--1327},
  year      = {2016}
}

@article{yin2026h,
  title   = {H-OPD: Confidence Aware Heterogeneous Multi-Teacher Multimodal On-policy Distillation},
  author  = {Yin, Qixiang and Yao, Huanjin and Cai, Yuchen and Chen, Jianghao and Wang, Ziyi and Yang, Min and Su, Fei and Zhao, Zhicheng},
  journal = {arXiv preprint arXiv:2607.02592},
  year    = {2026}
}

@inproceedings{wortsman2022model,
  title        = {Model soups: averaging weights of multiple fine-tuned models improves accuracy without increasing inference time},
  author       = {Wortsman, Mitchell and Ilharco, Gabriel and Gadre, Samir Ya and Roelofs, Rebecca and Gontijo-Lopes, Raphael and Morcos, Ari S and Namkoong, Hongseok and Farhadi, Ali and Carmon, Yair and Kornblith, Simon and others},
  booktitle    = {International conference on machine learning},
  pages        = {23965--23998},
  year         = {2022},
  organization = {Pmlr}
}

@article{bakouch2025smollm3,
  title   = {SmolLM3: smol, multilingual, long-context reasoner},
  author  = {Bakouch, Elie and Allal, Loubna Ben and Lozhkov, Anton and Tazi, Nouamane and Tunstall, Lewis and Pati{\~n}o, Carlos Miguel and Beeching, Edward and Roucher, Aymeric and Reedi, Aksel Joonas and Gallou{\'e}dec, Quentin and others},
  journal = {Hugging Face Blog},
  year    = {2025}
}
\bibliographystyle{iclr2027_conference}

\appendix

\section{Complete Method Details}
\label{app:method_details}

This appendix provides the complete capability controller, optimization
objective, and training procedure omitted from the main text for clarity.

\subsection{Capability Controller}
\label{app:ca_controller}

Recall that $T_d$ denotes the fixed calibrated capability of teacher $d$,
$C_d^{(k)}$ is the exponentially smoothed student capability estimate, and
\[
H_d^{(k)}=[T_d-C_d^{(k)}]_+
\]
denotes the remaining positive teacher advantage.

For domains with non-negligible initial headroom, the controller measures the
fraction of the initial capability deficit that remains:
\begin{equation}
r_d^{(k)}
=
\frac{H_d^{(k)}}{H_d^{(0)}},
\qquad
H_d^{(0)}>\epsilon.
\label{eq:app_headroom_standard}
\end{equation}

When the initial headroom is close to zero, directly dividing by
$H_d^{(0)}$ would make the controller sensitive to estimation noise.
We therefore use
\begin{equation}
r_d^{(k)}
=
\begin{cases}
H_d^{(k)}/H_d^{(0)}, &
H_d^{(0)}>\epsilon, \\[3pt]
0, &
H_d^{(0)}\le\epsilon
\ \text{and}\
H_d^{(k)}\le\epsilon, \\[3pt]
H_d^{(k)}/B_d^{(k)}, &
\text{otherwise},
\end{cases}
\label{eq:app_headroom_ratio}
\end{equation}
where
\begin{equation}
B_d^{(k)}
=
\max
\left\{
|T_d|,
|C_d^{\star(k)}|,
\epsilon
\right\}.
\label{eq:app_headroom_scale}
\end{equation}

The corresponding capability-headroom factor is
\begin{equation}
f_{d,\mathrm{head}}^{(k)}
=
\operatorname{clip}
\left(
\left(r_d^{(k)}\right)^p,
f_{\min},
f_{\max}
\right).
\label{eq:app_head_factor}
\end{equation}
We use $p=1$, $f_{\min}=0.05$, and $f_{\max}=2$.
The positive floor prevents a domain from being completely removed from
training when its estimated average capability deficit has closed.

\paragraph{Regression-aware retention.}
A small current teacher--student gap does not imply that the corresponding
capability will remain stable while other domains continue updating the shared
model. We therefore retain the historical-best student estimate
$C_d^{\star(k)}$ and define
\begin{equation}
R_d^{(k)}
=
\left[
C_d^{\star(k)}
-
C_d^{(k)}
-
\tau
\right]_+,
\label{eq:app_regression}
\end{equation}
where $\tau=0.02$.

When regression is detected, the retention factor is
\begin{equation}
f_{d,\mathrm{guard}}^{(k)}
=
\min
\left\{
f_{\max},
1+
\lambda
\frac{R_d^{(k)}}{B_d^{(k)}}
\right\},
\label{eq:app_guard}
\end{equation}
with $\lambda=1$.

The complete capability factor is
\begin{equation}
f_d^{(k)}
=
\begin{cases}
\max
\left\{
f_{d,\mathrm{head}}^{(k)},
f_{d,\mathrm{guard}}^{(k)}
\right\}, &
R_d^{(k)}>0, \\[3pt]
f_{d,\mathrm{head}}^{(k)}, &
R_d^{(k)}=0.
\end{cases}
\label{eq:app_capability_factor}
\end{equation}

We override the controller with $f_d^{(k)}=1$ during the first three
controller updates to avoid reacting immediately to noisy capability
estimates. The EMA coefficient is $\beta=0.9$, and the target token shares are
$q_d=1/3$ for the three domains used in our experiments.

The historical-best mechanism should be interpreted as a bounded response to
estimated regression rather than a hard non-forgetting guarantee. The
controller continuously adjusts supervision and does not permanently disable
or reactivate domains.

\subsection{Capability-Weighted OPD Objective}
\label{app:ca_objective}

All MOPD variants in our experiments use the same dense token-level teacher
signal and differ in how domain-level supervision is weighted.

For response position $t$ of example $i$, the candidate support
$\mathcal{S}_{it}$ consists of the old student's Top-$K$ predictions, with
$K=16$. The dense distillation signal follows Open-MOPD~\citep{gao2026open}:
\begin{equation}
A_{itv}
=
\operatorname{sg}
\left[
p_{itv}
\left(
\ell^{T}_{itv}
-
\ell^{\mathrm{old}}_{itv}
\right)
\right],
\qquad
v\in\mathcal{S}_{it}.
\label{eq:app_dense_advantage}
\end{equation}

CA-MOPD scales this signal using the effective domain multiplier
$w_{d(i)}^{(k)}$:
\begin{equation}
\widehat A_{itv}
=
\operatorname{sg}
\left[
w_{d(i)}^{(k)} A_{itv}
\right].
\label{eq:app_weighted_advantage}
\end{equation}

Let $s_d^{(k)}$ denote the observed valid-response-token share of domain $d$
in the current rollout batch. The token-balancing factor is
\begin{equation}
b_d^{(k)}
=
\frac{q_d}{s_d^{(k)}}.
\end{equation}

The final token multiplier is
\begin{equation}
w_d^{(k)}
=
F^{(k)}
\frac{
b_d^{(k)}f_d^{(k)}
}{
\sum_j
s_j^{(k)}
b_j^{(k)}
f_j^{(k)}
},
\qquad
F^{(k)}
=
\max_j f_j^{(k)}.
\label{eq:app_final_weight}
\end{equation}
This normalization satisfies
\begin{equation}
\sum_d s_d^{(k)} w_d^{(k)}
=
F^{(k)}.
\end{equation}

Thus, $f_d^{(k)}$ and $w_d^{(k)}$ represent different quantities.
The former captures capability-aware allocation, whereas the latter additionally
incorporates token-share balancing and global scaling. Likewise, an effective
weighted token share describes relative weighted-token contribution and should
not be interpreted as a direct measurement of gradient-vector magnitude or
parameter-space influence.

For optimization, let
\begin{equation}
\rho_{itv}(\theta)
=
\exp
\left(
\operatorname{clip}
\left(
\ell^\theta_{itv}
-
\ell^{\mathrm{old}}_{itv},
-20,
20
\right)
\right)
\label{eq:app_ratio}
\end{equation}
denote the candidate-level likelihood ratio.

We use the PPO clipped surrogate~\citep{schulman2017proximal} with the asymmetric clipping bounds of our Open-MOPD-based implementation~\citep{gao2026open}:
\begin{equation}
P_{itv}
=
\max
\left\{
-\widehat A_{itv}\rho_{itv},
-\widehat A_{itv}
\operatorname{clip}
\left(
\rho_{itv},
0.8,
1.28
\right)
\right\}.
\label{eq:app_ppo_clip}
\end{equation}

We apply dual clipping~\citep{ye2020mastering} with coefficient $c=3$, giving the candidate loss
\begin{equation}
\ell_{itv}
=
\begin{cases}
\min
\left\{
-3\widehat A_{itv},
P_{itv}
\right\}, &
\widehat A_{itv}<0, \\[3pt]
P_{itv}, &
\text{otherwise}.
\end{cases}
\label{eq:app_dual_clip}
\end{equation}

For an optimizer batch $\mathcal B$, the final token-normalized objective is
\begin{equation}
\mathcal L_{\mathrm{CA\text{-}MOPD}}(\theta)
=
\frac{
\sum_{i\in\mathcal B}
\sum_t
M_{it}
\sum_{v\in\mathcal S_{it}}
\ell_{itv}
}{
\sum_{i\in\mathcal B}
\sum_t
M_{it}
}.
\label{eq:app_objective}
\end{equation}

Entropy regularization and reference-policy KL regularization are disabled in
our main configuration. The dense OPD objective, candidate support, and
optimizer formulation are shared across the compared MOPD variants; CA-MOPD
changes the domain-level weighting mechanism rather than the underlying
token-level teacher signal.

\subsection{Training Algorithm}
\label{app:algorithm}

Algorithm~\ref{alg:camopd} summarizes the complete training procedure.

\begin{algorithmbox}{CA-MOPD training}
\label{alg:camopd}

\textbf{Input:}
shared checkpoint $\theta_0$;
specialists $\{\phi_d\}_{d\in\mathcal D}$;
calibration sets $\{\mathcal C_d\}$;
training prompt distributions $\{\mathcal P_d\}$.

\medskip
\textbf{1. Merge specialists.}

Construct
\[
    \theta_{\mathrm{merge}}
    =
    \theta_0
    +
    \sum_{d\in\mathcal D}
    (\phi_d-\theta_0),
\]
and initialize the student with
$\theta\leftarrow\theta_{\mathrm{merge}}$.

\medskip
\textbf{2. Calibrate capability.}

For each domain $d$:
\[
    T_d
    \leftarrow
    \widehat J_{\mathcal C_d}(\pi_{\phi_d}),
    \qquad
    C_d^{(0)}
    \leftarrow
    \widehat J_{\mathcal C_d}(\pi_\theta),
\]
\[
    C_d^{\star(0)}
    \leftarrow
    C_d^{(0)},
    \qquad
    H_d^{(0)}
    \leftarrow
    [T_d-C_d^{(0)}]_+.
\]

\medskip
\textbf{3. Repeat for each outer training step $k$.}

Sample prompts from all domains and generate student responses using
$\pi_{\theta_{\mathrm{old}}}$.

For each domain:
\begin{enumerate}
    \item aggregate verifier scores into $Z_d^{(k)}$;
    \item update
    \[
        C_d^{(k)}
        =
        \beta C_d^{(k-1)}
        +
        (1-\beta)Z_d^{(k)};
    \]
    \item update the historical best
    \[
        C_d^{\star(k)}
        =
        \max
        \left\{
            C_d^{\star(k-1)},
            C_d^{(k)}
        \right\};
    \]
    \item compute capability headroom and regression;
    \item obtain capability factor $f_d^{(k)}$;
    \item compute observed token share $s_d^{(k)}$ and effective multiplier
    $w_d^{(k)}$.
\end{enumerate}

Evaluate the corresponding teacher on the student's Top-$16$ candidate
tokens, construct $\widehat A_{itv}$, and optimize
Eq.~\eqref{eq:app_objective}.

\medskip
\textbf{Output:}
the final shared student $\pi_\theta$.

\end{algorithmbox}

\section{Implementation and Training Details}
\label{app:implementation}

\subsection{Training Configuration}
\label{app:training_config}

All experiments use the SmolLM3-3B model family. The mathematics,
code-generation, and instruction-following teachers are independently
RL-trained specialists derived from the same mixed-domain SFT checkpoint.
Table~\ref{tab:hyperparameters} summarizes the main training configuration.

\begin{table*}[t]
\centering
\small
\caption{
Training configuration used for the main CA-MOPD experiment.
}
\label{tab:hyperparameters}
\begin{tabular}{ll}
\toprule
Setting & Value \\
\midrule
Student / teachers
& SmolLM3-3B MixSFT / math, code, and IF RL specialists \\
Initialization
& Task Arithmetic, $\alpha_d=1$ \\
Hardware
& $2\times$ RTX PRO 5000 72GB \\
Precision
& BF16 \\
Training framework
& VERL / FSDP with vLLM rollout \\
Maximum prompt length
& 2,048 tokens \\
Maximum training response length
& 16,384 tokens \\
Prompts per outer step
& 12 (4 per domain) \\
Responses per prompt
& 4 \\
Training temperature
& 1.0 \\
Training top-$p$
& 0.99 \\
Training top-$k$
& $-1$ \\
Dense distillation support
& old-student Top-$16$ candidate tokens \\
Optimizer
& AdamW \\
Learning rate
& $1.5\times10^{-6}$ \\
Adam $\beta$
& $(0.9,0.999)$ \\
Weight decay
& 0.01 \\
Gradient clipping
& 1.0 \\
PPO clip range
& $(0.2,0.28)$ \\
Dual-clip coefficient
& 3 \\
Log-ratio clamp
& $[-20,20]$ \\
PPO epochs
& 1 \\
Mini-batch size
& 4 \\
Micro-batch size
& 1 \\
Loss aggregation
& token mean \\
Entropy regularization
& disabled \\
Reference KL
& disabled \\
Outer training steps
& 200 \\
Checkpoint steps
& 50, 100, 150, 200 \\
\bottomrule
\end{tabular}
\end{table*}

\subsection{Controller State and Checkpointing}
\label{app:checkpointing}

The capability controller is stateful. For each domain, we store
$C_d^{(k)}$, $C_d^{\star(k)}$, $H_d^{(0)}$, and the controller-update
counter. Resuming training therefore requires restoring the controller state
together with the model, optimizer, data-sampler, and rollout states.
Restoring model parameters alone can lead to a different subsequent weighting
trajectory even when the model checkpoint is identical.

Teacher references $T_d$ and initial student estimates $C_d^{(0)}$ are tied
to the corresponding calibration split and generation configuration. For
reproducibility, we record the calibration split identifier together with the
cached capability statistics.

\section{Capability Estimation and Evaluation Protocol}
\label{app:protocol}

\subsection{Capability Calibration}
\label{app:calibration}

For each domain $d$, we construct a calibration set
$\mathcal C_d$ containing 256 prompts. These prompts are excluded from
subsequent CA-MOPD training.

For each calibration prompt $x$, both the frozen teacher and the initial
student independently generate four responses using the same decoding
protocol. Let $y_{x,1},\ldots,y_{x,4}$ denote these responses. We first
compute the prompt-level capability estimate
\begin{equation}
\widehat J_x(\pi)
=
\frac{1}{4}
\sum_{r=1}^{4}
V_d(x,y_{x,r}),
\end{equation}
and then average equally across prompts:
\begin{equation}
\widehat J_{\mathcal C_d}(\pi)
=
\frac{1}{|\mathcal C_d|}
\sum_{x\in\mathcal C_d}
\widehat J_x(\pi).
\end{equation}

This two-stage aggregation gives equal weight to every prompt regardless of
the number of sampled responses.

The teacher reference
\begin{equation}
T_d = \widehat J_{\mathcal C_d}(\pi_{\phi_d})
\end{equation}
is computed once and remains fixed throughout training. The initial student
capability
\begin{equation}
C_d^{(0)} = \widehat J_{\mathcal C_d}(\pi_{\theta_{\mathrm{merge}}})
\end{equation}
initializes both the EMA state and the historical-best state.

During subsequent training, no additional student rollout is generated solely
for capability estimation. Instead, the controller reuses verifier outcomes
from responses already produced by the on-policy rollout stage.

Calibration scores and online capability estimates should therefore be
interpreted as operational capability estimates under the specified prompt
distribution and verifier, rather than as claims about task performance
outside that evaluation setting.

\subsection{Benchmark Evaluation}
\label{app:ca_eval}

We evaluate mathematics using AIME24 and AIME25, code generation using
LiveCodeBench v5 and v6, and instruction following using IFEval and IFBench.
Table~\ref{tab:eval_protocol} summarizes the benchmark-specific evaluation
configuration.

\begin{table}[t]
\centering
\small
\setlength{\tabcolsep}{3.5pt}
\caption{
Benchmark-specific inference and evaluation settings.
All evaluations use thinking mode.
Top-$k=-1$ disables top-$k$ filtering.
}
\label{tab:eval_protocol}
\begin{tabular}{llrrrrrl}
\toprule
Domain & Benchmark & Temp. & Top-$p$ & Top-$k$
& \shortstack{Max Output\\Tokens}
& \shortstack{Max Sequence\\Length}
& Metric \\
\midrule
Math & AIME24
& 0.6 & 0.95 & $-1$ & 30,000 & 32,768 & avg@64 \\

Math & AIME25
& 0.6 & 0.95 & $-1$ & 30,000 & 32,768 & avg@64 \\

Code & LCB v5
& 1.0 & 0.95 & $-1$ & 30,000 & 32,768 & avg@10 \\

Code & LCB v6
& 1.0 & 0.95 & $-1$ & 30,000 & 32,768 & avg@10 \\

IF & IFEval
& 1.0 & 0.95 & $-1$ & 30,000 & 32,768 & strict prompt \\

IF & IFBench
& 1.0 & 0.95 & $-1$ & 30,000 & 32,768 & strict prompt \\
\bottomrule
\end{tabular}
\end{table}

For AIME, we compute the verifier score of each independently generated
response, average these scores within each problem, and then average equally
across problems. We therefore report average sampled accuracy rather than
majority-vote accuracy or pass@$k$.

For LiveCodeBench, we independently sample ten programs for each problem and evaluate each program against the benchmark tests. We first average the binary correctness scores over the ten samples for each problem, and then average equally across problems. We report this metric as avg@10.

IFEval and IFBench use strict prompt-level correctness: a prompt is counted as
successful only when all required instructions are satisfied.

\end{document}